\documentclass[10pt,twocolumn,letterpaper]{article}

\usepackage[pagenumbers]{cvpr} 

\usepackage{graphicx}
\usepackage{amsmath}
\usepackage{amssymb}
\usepackage{booktabs}
\usepackage{multirow}
\usepackage{verbatim}
\usepackage[accsupp]{axessibility}

\usepackage[pagebackref,breaklinks,colorlinks]{hyperref}

\usepackage[capitalize]{cleveref}
\crefname{section}{Sec.}{Secs.}
\Crefname{section}{Section}{Sections}
\Crefname{table}{Table}{Tables}
\crefname{table}{Tab.}{Tabs.}

\def\cvprPaperID{5677} 
\def\confName{CVPR}
\def\confYear{2023}

\begin{document}

\title{LINe: Out-of-Distribution Detection by Leveraging Important Neurons}

\author{Yong Hyun Ahn$^1$, Gyeong-Moon Park$^{2,}$\textsuperscript{*}, Seong Tae Kim$^{2,}$\thanks{Corresponding authors: Gyeong-Moon Park (gmpark@khu.ac.kr) and Seong Tae
Kim (st.kim@khu.ac.kr).}\\
$^1$Department of Artificial Intelligence, Kyung Hee University \\ 
$^2$Department of Computer Science and Engineering, Kyung Hee University\\
}
\maketitle

\begin{abstract}
 It is important to quantify the uncertainty of input samples, especially in mission-critical domains such as autonomous driving and healthcare, where failure predictions on out-of-distribution (OOD) data are likely to cause big problems. OOD detection problem fundamentally begins in that the model cannot express what it is not aware of. Post-hoc OOD detection approaches are widely explored because they do not require an additional re-training process which might degrade the model's performance and increase the training cost.
 In this study, from the perspective of neurons in the deep layer of the model representing high-level features, we introduce a new aspect for analyzing the difference in model outputs between in-distribution data and OOD data. We propose a novel method, Leveraging Important Neurons (LINe), for post-hoc Out of distribution detection.
 Shapley value-based pruning reduces the effects of noisy outputs by selecting only high-contribution neurons for predicting specific classes of input data and masking the rest.
 Activation clipping fixes all values above a certain threshold into the same value, allowing LINe to treat all the class-specific features equally and just consider the difference between the number of activated feature differences between in-distribution and OOD data.
 Comprehensive experiments verify the effectiveness of the proposed method by outperforming state-of-the-art post-hoc OOD detection methods on CIFAR-10, CIFAR-100, and ImageNet datasets. Code is available on \href{https://github.com/YongHyun-Ahn/LINe-Out-of-Distribution-Detection-by-Leveraging-Important-Neurons}{https://github.com/LINe-OOD}

\end{abstract}

\section{Introduction}
\label{sec:intro}

Recently, deep learning has made tremendous advances in various fields.
This advancement has captivated numerous researchers, leading to many attempts to apply deep learning techniques to real-world applications. 
However, applying these state-of-the-art techniques to real-world applications is often limited for several reasons.
One primary obstacle is the presence of unseen classes of samples during training. 
These samples, referred to as out-of-distribution (OOD) data, can compromise a model's stability and, in some cases, severely impair its performance. 
The inherent characteristics of OOD samples can lead to potentially severe consequences in mission-critical domains, such as autonomous driving and medical applications.
So, effectively handling these OOD samples is vital to avoid problems, such as car crashes and misdiagnosis.

\begin{figure}[t]
    \centering
    \includegraphics[width=\linewidth]{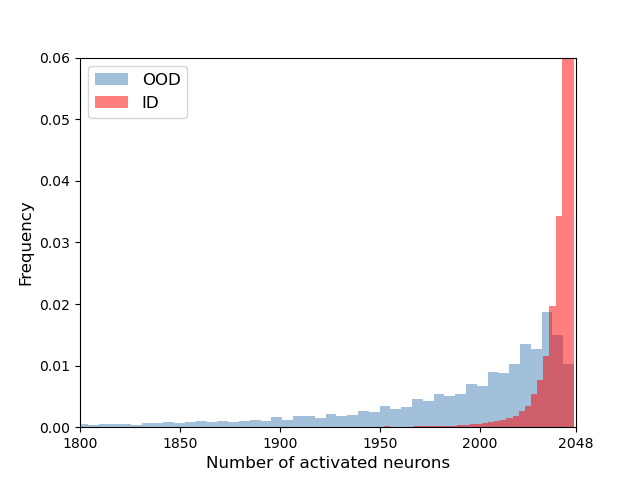}
    \caption{\small\textbf{Illustration of the number of activated neuron in the penultimate layer.} The distributions of the number of activated neurons from ID (ImageNet\cite{huang2021mos}) and two OOD datasets (textures\cite{cimpoi2014describing} and iNaturalist\cite{inat}) are presented.
    We define an activated neuron as one fired with an activation value greater than zero.
    Most of the neurons are activated when the model receives ID samples but fewer neurons are activated in OOD samples.
    }
    \label{fig:whyact}
\end{figure}

For OOD detection, numerous techniques have been explored to analyze the distinction between in-distribution (ID) data and OOD data. 
There are several approaches to OOD detection, including confidence-based methods \cite{liang2017enhancing,devries2018learning,bendale2016towards,Kevin,huang2021mos}, density-based methods \cite{Lee,zong2018deep,abati2019latent,pidhorskyi2018generative,sabokrou2018adversarially,kobyzev2020normalizing,zisselman2020deep,kingma2018glow,van2016pixel,jiang2021revisiting,nalisnick2018deep,kirichenko2020normalizing}, and distance-based methods\cite{lee2018simple,sun2022out,ren2021simple,techapanurak2020hyperparameter,chen2020boundary,zaeemzadeh2021out,van2020uncertainty,huang2020feature,ming2022cider}.
The post-hoc method is one approach in OOD detection that offers significant advantages in real-world applications as it eliminates the need for a re-training process, which could potentially degrade the model's performance and increase training costs \cite{yang2021generalized}.

Post-hoc OOD detection methods  \cite{liu2020energy, sun2021react, sun2022dice} employ outputs such as model logits or layer activations, which are typically used for prediction, to calculate OOD scores. 
These scores allow for the differentiation of ID and OOD data based on the disparities in their respective scores \cite{yang2021generalized}. 
Enhancing the overall OOD score distribution difference between ID and OOD data is a critical aspect in improving the performance of post-hoc OOD detection methods.
Recent studies \cite{sun2021react, sun2022dice} have found that augmenting the overall difference is contingent upon the capacity to mitigate noisy signals. 
For example, ReAct \cite{sun2021react} demonstrates that OOD data exhibit considerable values in the penultimate layer activation. 
By truncating these noisy activations, ReAct effectively improves OOD detection performance. 
Similarly, DICE \cite{sun2022dice} uncovers the presence of noisy signals that increase the variance of the OOD score distribution. 
By selectively employing the most salient weights, DICE enhances OOD detection performance while reducing the impact of noisy signals.
Minimizing the influence of noisy model outputs is crucial for advancing post-hoc OOD detection performance. 
In this study, we reveal an additional factor that is also important in the context of OOD detection.

Figure \ref{fig:whyact} displays the histogram of the number of activated features in the penultimate layer. 
We define neurons with activation values greater than zero as activated neurons. 
As the figure demonstrates, a majority of neurons are activated when the model encounters ID samples. 
However, for OOD samples, fewer neurons are activated.
To understand the primary cause of the observation in Figure \ref{fig:whyact}, we investigate the model from the perspective of neuron-concept association \cite{bau2017dissection,bau2020understanding,wang2022hint,khakzar2021neural}. 
According to \cite{bau2017dissection}, neurons are trained to detect disentangled high-level features in the deep layers of convolutional neural networks (CNNs), and they can even learn new unlabeled abstract concepts from the data \cite{wang2022hint}. 
The number of neurons representing these high-level features increases in deeper layers and becomes the most significant amount in the penultimate layer \cite{bau2020understanding}.
Since the activation in the penultimate layer represents the presence of these high-level features \cite{bau2017dissection}, different input images activate high-level features in varying ways, such as magnitude and patterns. 
These distinct patterns of activation in the penultimate layer are ultimately used to predict the input image's class. 
As each class possesses unique characteristics related to high-level concepts, the associated high-level features differ for each class. Consequently, each associated high-level feature can be categorized into one class-specific feature group.
The disparity in associated high-level features in neurons results in a difference in penultimate layer activation between ID and OOD samples, which are predicted as the same class. 
Thus, taking into account the number of activated essential neurons can serve as a useful indicator for distinguishing ID and OOD samples.

In this paper, we present a novel post-hoc OOD detection method called \emph{\textbf{L}everaging \textbf{I}mportant \textbf{Ne}urons} (\textbf{LINe}). 
LINe harnesses two crucial aspects: considering the number of activated important neurons and minimizing noisy activations to enhance OOD detection performance. 
To achieve this, we introduce two powerful techniques within LINe: 1) Shapley-based pruning and 2) activation clipping (AC).

Shapley value-based pruning is a method that mitigates the influence of noisy outputs by selecting activations essential for inferring input data classes. 
While several methods exist for identifying important activations, Sun et al. \cite{sun2022dice} use activation magnitude to determine their importance. 
However, this approach alone is insufficient for quantifying a neuron's importance. 
Hence, we employ the Shapley value \cite{shapley1997value} to more accurately measure each neuron's contribution. 
Applying the Shapley value concept \cite{shapley1997value} to neural networks allows us to quantify each neuron's contribution to identifying a specific class.
Moreover, neurons with high Shapley values are associated with critical input image features \cite{khakzar2021neural}. 
By leveraging the Shapley value \cite{shapley1997value}, we can identify neurons that represent important high-level features for each class, which are essential for reducing noisy output.

Activation clipping, a concept introduced in ReAct \cite{sun2021react}, is another powerful technique. 
Interpreting activation clipping in terms of class-specific features provides a new understanding of its role in considering the number of activated important neurons for OOD detection. 
Activation clipping adjusts values exceeding a certain threshold to the threshold value. 
This modification enables us to account for the differences in the number of activated features between in-distribution and OOD data by treating numerous class-specific features equally. 
By considering the variation in the number of activated class-specific features, we can effectively augment the overall OOD score difference between ID and OOD data, leading to enhanced performance.

Our key contributions are summarized as follows:
\begin{itemize}
    \item 
    We propose a simple yet effective post-hoc OOD detection method, named LINe, which uses the Shapley value to rank the contribution of neurons and gives a new inspiration for leveraging selected important class-specific neurons.
    \item 
    We unveil the important factor for improving OOD scoring and show a new way of understanding the role of activation clipping for OOD detection. By applying activation clipping, we can fully consider the number of activated class-specific features and achieve higher OOD detection performance.
    
    \item
    Comprehensive experiments have been conducted to verify the effectiveness of the proposed method on the CIFAR-10, CIFAR-100, and ImageNet-1K. Compared to the competitive post-hoc method DICE\cite{sun2022dice}, LINe reduces the FPR95 by up to 14.05$\%$.
\end{itemize}

\section{Background and Related Work}
\label{sec:Related}

\subsection{Neuron-Concept Association}
\label{sec:Related-NC associate}
Neuron-concept association methods are a field of study that tries to interpret the internal computation of CNN to a human-understandable concept \cite{mu2020compositional,chen2019looks,bargal2018guided,graziani2018regression}. 
Several studies have shown that neurons of shallower layers tend to learn more simple and low-level concepts, such as curves and edges, while deeper layers learn more abstract and high-level concepts, such as arm and face\cite{zhou2018interpreting,zhou2014object,wang2022hint}.
The methods for quantifying the concept's contribution are also introduced in \cite{lu2020geometry,ghorbani2019towards,graziani2018regression}.
Network Dissection\cite{bau2017dissection,bau2020understanding,zhou2018interpreting} assigns each neuron to a concept to quantify its role.
Bau et al. \cite{bau2019gan} investigate the effect of concept-specific neurons by observing the change of concept-related contents in generative models.
Recently, Wang et al.  \cite{wang2022hint} show that models can learn abstract concepts like mammal and carnivore, which is not in the label set of training data.

\subsection{Shapley Value}
\label{sec:Related-Shap}
Shapely value\cite{shapley1997value,kuhn1953contributions,sundararajan2020many} is a concept from Game Theory, which evaluates each property's individual and collaborative effects. 
Studies have been conducted using Shapley value in CNNs to measure each neuron's contribution and interpret models' behaviors \cite{lundberg2017unified,ancona2019explaining,ghorbani2020neuron,sundararajan2020many,khakzar2021neural}.
Neuron Shapley\cite{ghorbani2020neuron} sorts the Shapley values to identify the most influential neurons from all hidden layers as image categories.
Khazar et al. \cite{khakzar2021neural} show neurons that have high Shapley values have high correlations with important features of the input image.
Previous studies have inspired us to select important class-specific neurons through the contribution scores calculated from the Shapley value.
LINe can effectively eliminate the negative effects of noisy signals by leveraging the selected class-specific neurons.

\subsection{Out-of-Distribution Detection}
\label{sec:Related-OOD}
OOD detection aims to find inputs with different characteristics from the training data \cite{yang2021generalized}. 
A lot of research efforts have been devoted to developing an effective method to distinguish OOD inputs from ID inputs.  
Confidence-based methods perform OOD detection by quantifying OOD scores based on different scoring functions \cite{liang2017enhancing,devries2018learning,bendale2016towards,Kevin,huang2021mos}.
Hendrycks et al. \cite{Kevin} used a maximum softmax probability (MSP) of the model as a baseline confidence-base OOD scoring function. ODIN\cite{liang2018enhancing} utilizes perturbation of inputs and temperature scaling on the softmax layer to increase the difference between ID and OOD. 
To enhance the effectiveness of confidence-based scores, recently, Liu et al. \cite{liu2020energy} introduced an energy-based score with the theoretical interpretation from a likelihood perspective, which is further adopted in \cite{lin2021mood,wang2021can,morteza2022provable} to distinguish ID and OOD samples.
Distance-based approaches measure the distance between input sample and typical ID samples or centroids of them \cite{lee2018simple,sun2022out,ren2021simple,techapanurak2020hyperparameter,chen2020boundary,zaeemzadeh2021out,van2020uncertainty,huang2020feature,ming2022cider}.
These approaches are based on simple evidence that OOD samples should have more distance than IDs. 
Similarly, density-based methods identify OOD samples based on the distribution of the training samples and use density (or likelihood) \cite{Lee,zong2018deep,abati2019latent,pidhorskyi2018generative,sabokrou2018adversarially,kobyzev2020normalizing,zisselman2020deep,kingma2018glow,van2016pixel,jiang2021revisiting,nalisnick2018deep,kirichenko2020normalizing}.

But none of the aforementioned methods consider the number of activated features, which can be a good indicator to distinguish ID and OOD samples.
The most similar study to our study is DICE\cite{sun2022dice}.
DICE leverages sparsification to reduce the effect of noisy signals by selectively using salient weights from activation \cite{sun2022dice}. 
In this study, we effectively eliminate the outcomes of noisy signals by accurately selecting neurons leveraging the contribution of neurons calculated from Shapley value\cite{shapley1997value,khakzar2021neural}.
In addition, our method performs OOD detection more effectively by considering the number of activated features. 
The neurons are known to be associated with the concepts, and the activation patterns of neurons in deep layers are different in ID and OOD.
This gives a theoretical background for considering the number of activation in OOD detection. 
To the best of our knowledge, our work is the first study to leverage the neuron contribution based on Shapley value and consider the number of activated features for calculating OOD score. 

\section{Method}
\label{sec:Method}
\subsection{Method Overview}

\begin{figure*}[t]
    \centering
    \includegraphics[width=\textwidth]{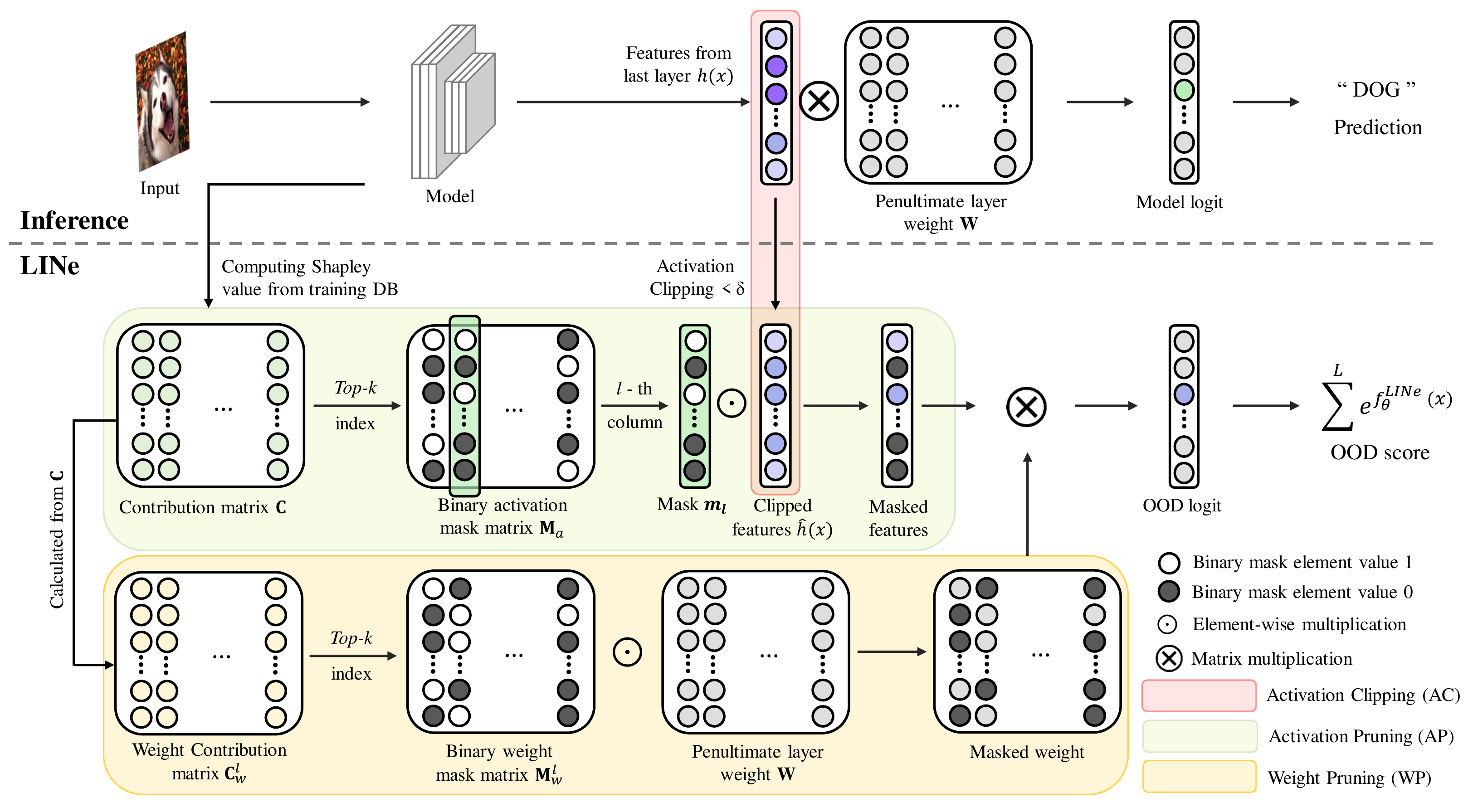}
    \caption{\small\textbf{Overall procedure of LINe for calculating OOD score.} LINe is composed of mainly three parts. 1) Activation Clipping (AC), 2) Activation Pruning (AP), 3) Weight Pruning (WP).
    AC limits the magnitude of the activation, which allows LINe to consider the number of activated features.
    AP uses contribution matrix $\mathbf{C}$, which is precomputed from Shapley value\cite{shapley1997value}. AP leverages $\mathbf{C}$ to select important neurons and mask others.
    WP uses an adjusted contribution matrix which is obtained from the contribution matrix $\mathbf{C}$. Leveraging AP and WP, LINe can reduce noisy signals effectively.
    }
    \label{fig:overall_concept}
\end{figure*}

Our method mainly consists of two parts: \textit{Activation Clipping} and \textit{Shapley-based pruning}.
Activation clipping is a method of clipping each neuron activation to a specific value when the activation value exceeds a particular threshold ($\delta$).
Through activation clipping, our method can consider the number of high-level features during calculating OOD scores, which are analyzed in terms of neuron-concept association. 
Next, Shapley-based pruning eliminates the negative effect of noisy signals by measuring neural contributions of neural networks using important neurons only.
We can accurately measure the contribution of each neuron using Shapley value, which is a mathematically grounded method.
Also, by applying Shapley value, we can obtain supporting evidence that neurons with large contributions represent critical features for recognizing input image\cite{khakzar2021neural}.
This allows us to assemble all contributions for each class and select important class-specific neurons representing class-specific features.
More details of activation clipping and Shapley-based pruning will be described in Subsection \ref{sec:AC} and Subsection \ref{sec:pruning}.
In Subsection \ref{sec:LINe}, we will explain the overall method. 

\subsection{Activation Clipping}
\label{sec:AC}
For a pre-trained deep neural network $f_{\theta}(x)$ parameterized by $\theta$, $f_{\theta}(x)$ encodes an input $\*x \in \mathbb{R}^{d}$, where $d$ indicates dimension of input $\*x$, and predicts a class distribution for $L$ different classes, i.e., $f_\theta(\*x)\in \mathbb{R}^{L}$.
Feature vector from the penultimate layer of the network denotes $ h(\*x) \in \mathbb{R}^q$, where $q$ stands for the dimension of penultimate layer output $h(\*x)$.
Weight matrix $\mathbf{W} \in \mathbb{R}^{q\times L}$ weights the importance of each feature in $h(\*x)$ and transfers to output $f_\theta(\*x)$ as follows:

\begin{equation}\label{equ:formulation}
f_{\theta}(x) = \textbf{W}^{T} h(x) + \textbf{b}.
\end{equation}

Activation clipping is applied to the feature vector in the penultimate layer. 
For each neuron that is activated above a certain threshold, AC limits the magnitude of the activation. 
As we discussed in Figure \ref{fig:whyact}, the number of activated features in ID and OOD samples are different.
By limiting the magnitude of the activation to the same value $\delta$, we can treat every activated high-level feature equally, which allows us to consider the number of activated features in the OOD score.
This increases the OOD score difference between ID and OOD samples, thereby improving performance.

For each activation $a_{i}$ in $h(x)$, penultimate layer feature $h(x)$ can be denoted as $h(x) = \left[a_{1}, a_{2}, \cdots,a_{q} \right]$. 
With a clipping threshold $\delta$, the clipped activation $\hat{a}_{i}$ can be described as  $\hat{a}_{i} = min(a_{i},\delta)$. 
The clipped feature vector $\hat{h}(\*x)$ is then,
\begin{equation}\label{clip}
    \hat{h}(\*x) = \left[\hat{a}_{1}, \hat{a}_{2}, \cdots,\hat{a}_{q} \right].
\end{equation}
The model output after AC can be given as:
\begin{equation}\label{equ:clip}
    f_{\theta}^{AC}(x) = \textbf{W}^{T} \hat{h}(x) + \textbf{b}.
\end{equation}

\subsection{Shapley-based Pruning}
\label{sec:pruning}

Shapley-based pruning selectively uses a subset of important neuron activation and weights. 
To select these subsets, we calculate Shapley value\cite{shapley1997value} defined by an average of the effect of removing a single unit (e.g., marginal contribution) to all possible combinations of units. 
However, computing all the combinations of units is computationally expensive and practically infeasible for recent large neural networks.
Therefore, we use the Taylor approximation to compute the Shapley value, which is introduced in \cite{khakzar2021neural}.
For input $x^{l} \in \mathit{D}$, where $x^{l}$ denotes the sample of class $l$ from dataset $\mathit{D}$, a contribution(i.e., Shapley value) of $i$-th neuron $a_{i}$ in class $l$, $s_{i}^{l}$ is calculated as 

\begin{equation}\label{equ:taylor}
    s_{i}^{l} = \left| f_{\theta}(x^{l}) - f_{\theta}(x^{l};a_{i}\leftarrow 0)\right| = \left| a_{i}\nabla_{a_{i}}f_{\theta}(x^{l}) \right|.
\end{equation}

\subsubsection{Activation Pruning}
\label{sec:AP}
Activation pruning (AP) selectively uses the subset of important neuron activation.
Through AP, We can effectively reduce the impact of noisy activation.
From the contribution of each neuron $s_{i}^{l}$, which is using obtained from all training data,
contribution matrix $\mathbf{C} \in \mathbb{R}^{q\times L}$ is defined as the class-specific average of all contribution $s_{i}^{l}$.
An ($i$,$l$)-th entry of contribution matrix $c_{il} \in \mathbf{C}$  is defined as:
\begin{equation}\label{equ:cont-component}
    \textbf{C}_{il} = \frac{1}{n}\sum^{n}s_{i}^{l},
\end{equation}
where $n$ denotes the number of training images in class $l$.
\begin{figure}[t]
    \centering
    \includegraphics[width=\linewidth]{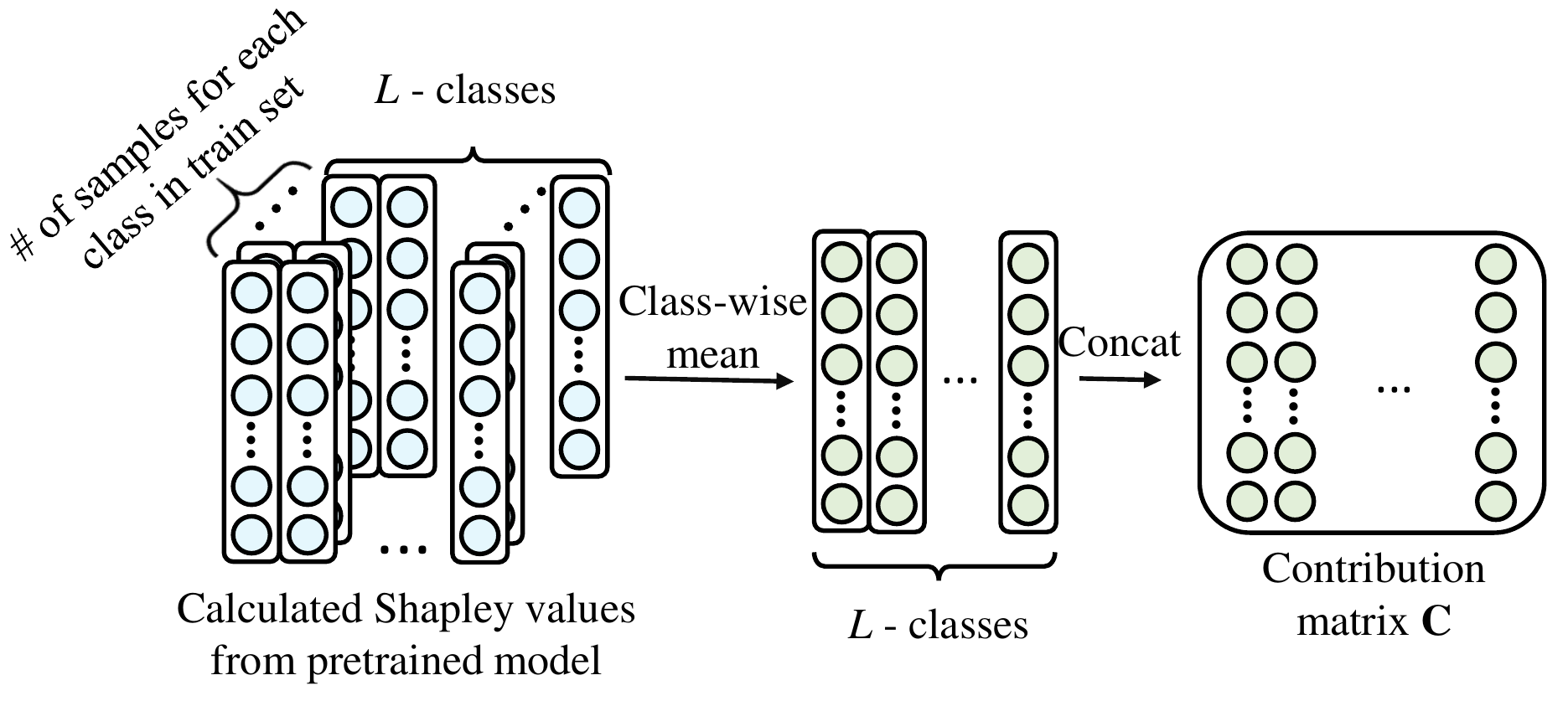}
    \caption{\small\textbf{Calculation of contribution matrix $\mathbf{C}$.} 
    }
    \label{fig:c}
\end{figure}
We select \emph{top-k} neurons for each class based on the k-largest elements from each column in $\mathbf{C}$ and define an activation mask matrix $\mathbf{M}_{a} \in \mathbb{R}^{q\times L}$, where we set 1 for the $k$-largest elements from every column in $\mathbf{C}$, otherwise 0.
The model output after AP with the predicted class $l$ is given as:
\begin{equation}\label{equ:AP}
    f_{\theta}^{AP}(x) = \textbf{W}^{T} (\mathbf{m}_{l} \odot h(x)) + \textbf{b},
\end{equation}
where $\mathbf{m}_{l} \in \mathbb{R}^q$ indicates $l$-th column of mask matrix $\mathbf{M}_{a}$ and $\odot$ denotes the element-wise multiplication.

\subsubsection{Weight Pruning}
\label{sec:WP}
Weight pruning (WP) selectively uses the subset of important penultimate layer weights.
Through WP, We can effectively reduce the impact of noisy signals due to the overparameterized model.
The contribution of neurons is further used to refine the weight matrix as WP. 
For this purpose, we define a weight contribution matrix for the class $l$ as $\mathbf{C}_{w}^{l} = \mathbf{c}_{l} \odot \mathbf{W} \in \mathbb{R}^{q\times L}$ where $\mathbf{c_{l}}$ indicates the $l$-th column of contribution matrix $\mathbf{C}$. 
We select \emph{top-k} weights for each class based on the $k$-largest elements in $\mathbf{C}_{w}^{l}$ and define a mask matrix for class $l$ as $\mathbf{M}_{w}^{l} \in \mathbb{R}^{q\times L}$ by setting 1 for the k-largest elements in $\mathbf{C}_{w}^{l}$, otherwise 0.
The model output after WP with the predicted class $l$ is given as:
\begin{equation}\label{equ:WP}
    f_{\theta}^{WP}(x) = (\textbf{W} \odot \textbf{M}_{w}^{l})^{T} h(x) + \textbf{b}.
\end{equation}
From the Shapley-based AP and WP, we can determine the neurons representing important high-level features for each class, which play a crucial role in reducing noisy output.\\

\subsection{Leveraging Important Neurons (LINe)}
\label{sec:LINe}
As already described in the previous Subsection \ref{sec:AC} and \ref{sec:pruning}, there are two ideas to improve the performance of post-hoc OOD detection.
One is considering the number of activated high-level features, and the other is reducing noisy signals from less important neurons.
LINe achieves both ways to improve post-hoc OOD detection performance through the procedures described above.
By adapting LINe to a model, we can effectively increase the overall difference in OOD scores between ID and OOD data.
As a result, the model output under LINe with the predicted class $l$ is described as 
\begin{equation}\label{equ:LINe}
    f_{\theta}^{LINe}(x) = (\textbf{W} \odot \textbf{M}_{w}^{l})^{T}(\mathbf{m}_{l} \odot \hat{h}(x)) + \textbf{b}.
\end{equation}
Please note that there is no parameter change in the model itself and ID classification accuracy can be preserved.

\begin{figure}[t]
    \centering
    \includegraphics[width=\linewidth]{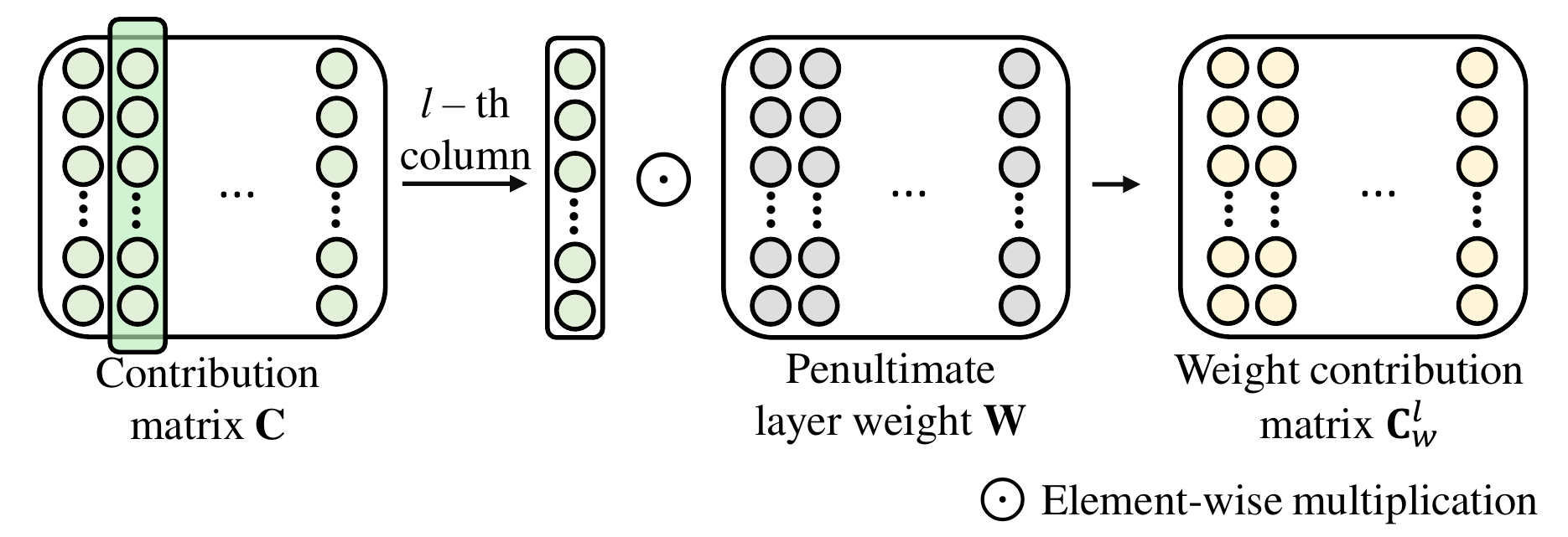}
    \caption{\small\textbf{Calculation of contribution matrix $\mathbf{C_{w}^{l}}$.} 
    }
    \label{fig:c}
\end{figure}

\section{Experiments}
\label{sec:Experiments}

Comprehensive experiments have been conducted to evaluate our method.
In section 4.1, we used the CIFAR\cite{krizhevsky2009learning} benchmark, which is one of the famous benchmarks in OOD studies\cite{sun2021react,liu2020energy,sun2022dice}. 
In Section 4.2, experiments were conducted based on a large-scale dataset, ImageNet, with various OOD datasets.
Section 4.3 analyzes why our method is effective through various ablation experiments.

\subsection{Evaluation on CIFAR Benchmarks}
\noindent \textbf{Implementation Details.}
\label{sec:imp-cifar}
In this experiment, we used 10,000 test images each from CIFAR-10\cite{krizhevsky2009learning} and CIFAR-100\cite{krizhevsky2009learning} as ID data, respectively.
The model's performance was evaluated using six commonly used OOD datasets as an OOD benchmark.
The list of six OOD datasets is as follows:
SVHN\cite{netzer2011reading}, Textures\cite{cimpoi2014describing},  iSUN\cite{xu2015turkergaze}, 
LSUN-Crop\cite{yu2015lsun}, 
LSUN-Resize\cite{yu2015lsun}, and Places365\cite{zhou2017places}.
As the pre-trained models, we used DenseNet\cite{huang2017densely}. As in \cite{sun2022dice}, the models are trained from CIFAR-10\cite{krizhevsky2009learning} and CIFAR-100\cite{krizhevsky2009learning} with 50,000 training images respectively.
Following \cite{sun2022dice}, the models were trained during 100 epochs with batch size 64, weight decay 0.0001, momentum 0.9, and start learning rate 0.1. The learning rate was decayed by a factor of 10 at epochs 50, 75, and 90.
We used the entire training dataset to estimate the contribution matrix $C$.

\begin{table}[t]
\caption{\textbf{Comparison on CIFAR benchmarks.} Comparison with competitive post-hoc OOD detection methods on CIFAR benchmarks. All values in this table are percentages and averaged over six OOD test datasets.
$\downarrow$ indicates smaller value means better performance and $\uparrow$ indicates vice versa. \textbf{Bold} numbers are superior results.
The overall detail results for each OOD dataset are provided in supplementary material.
}

\scalebox{0.8}{
\begin{tabular}{lcc|cc}
\toprule
\multirow{2}{*}{\textbf{Method}} & \multicolumn{2}{c|}{\textbf{CIFAR-10}} & \multicolumn{2}{c}{\textbf{CIFAR-100}} \\
 \multicolumn{1}{c}{} & \multicolumn{1}{c}{\textbf{FPR95 $\downarrow$}} & \multicolumn{1}{c|}{\textbf{AUROC $\uparrow$}} & \multicolumn{1}{c}{\textbf{FPR95 $\downarrow$}} & \multicolumn{1}{c}{\textbf{AUROC $\uparrow$}}\\
 \midrule
 MSP\cite{Kevin}  & 48.73 & 92.46 & 80.13 & 74.36 \\
 ODIN\cite{liang2018enhancing}  & 24.57 & 93.71 & 58.14 & 84.49  \\
 Mahalanobis\cite{lee2018simple}  & 31.42	 & 89.15 & 55.37 &	82.73 \\
 Energy\cite{liu2020energy}  & 26.55 & 94.57 & 68.45 & 81.19  \\
 ReAct\cite{sun2021react}  & 26.45	& 94.95 & 62.27 & 84.47  \\
DICE\cite{sun2022dice} & 20.83 & 95.24 & 49.72 & 87.23 \\
DICE + ReAct & 16.48 & 96.64 & 49.57 & 85.08 \\
\textbf{LINe (Ours)} & \textbf{14.71} & \textbf{96.99} & \textbf{35.67} & \textbf{88.67} \\ 
\bottomrule
\end{tabular}}
\label{tab:cifar-results}
\end{table}

\noindent \textbf{Comparison.}
For comparison, we adopted recent post-hoc OOD detection methods:
MSP \cite{Kevin},
ODIN\cite{liang2018enhancing}, 
Mahalanobis distance\cite{lee2018simple}, 
Energy \cite{liu2020energy}, 
ReAct\cite{sun2021react}, 
and DICE\cite{sun2022dice}. 

For all methods, the performances were measured by OOD scores, derived from the same DenseNet model.

\noindent \textbf{Experimental Results.}
Table \ref{tab:cifar-results} shows the comparisons between LINe and other post-hoc OOD detection methods on CIFAR-10 and CIFAR-100 benchmarks. 
As shown in the table, our method achieved state-of-the-art performances by outperforming all the other methods on both CIFAR-10 and CIFAR-100 datasets. 
In CIFAR-100, LINe reduced FPR 95 by 14.05$\%$ compared to the competitive method DICE\cite{sun2022dice}. 
In CIFAR-100, LINe achieved an FPR 95 of 14.05$\%$ which was lower than the FPR 95 of DICE\cite{sun2022dice} (49.72\%) and DICE + ReAct (49.57\%).
DICE + ReAct is the method that is implemented by applying ReAct \cite{sun2021react} on DICE \cite{sun2022dice}.
DICE\cite{sun2022dice} removed the noisy signals by using the magnitude of activation and weight.
LINe not only removed the noisy signals using class-specific neurons but also considered the number of activations of the high-level feature.

\subsection{Evaluation on ImageNet}
\label{sec:imagenet}

\noindent \textbf{Implementation Details.} 
In real-world applications, the model encounters high-resolution images with various scenes and features, and evaluation on a large-scale dataset can provide clues about model performance in a real-world application.
Therefore, in this experiment, we evaluated LINe on a large-scale ImageNet dataset.
Based on \cite{huang2021mos}, a subset of the four datasets where all the overlapping categories with ImageNet-1k were eliminated was used as OOD datasets. 
The four OOD datasets are as follows:
Textures\cite{cimpoi2014describing},
Places365\cite{zhou2017places},
iNaturalist\cite{inat},
and SUN\cite{sun}.
We used a pre-trained ResNet-50 model\cite{he2016identity}, which is trained with ImageNet-1k.
The entire training dataset was used to estimate the contribution matrix $C$, and all images were resized to 224 $\times$ 224 at test time.

\begin{table*}[t]
\centering
\caption{\textbf{Comparison on ImageNet benchmark.} Comparisons with competitive post-hoc OOD detection methods on ImageNet benchmark. All values in this table are percentages and averaged over four OOD test datasets.
} 
\label{tab:imagenet-results}
\scalebox{0.83}{
\begin{tabular}{lcccccccccc}
    \toprule
  \multirow{3}{*}{\textbf{Method}}  & \multicolumn{8}{c}{\textbf{OOD Datasets}} & \multicolumn{2}{c}{\multirow{2}{*}{\textbf{Average}}} \\ \cline{2-9}
 \multicolumn{1}{c}{} & \multicolumn{2}{c}{\textbf{iNaturalist}} & \multicolumn{2}{c}{\textbf{SUN}} & \multicolumn{2}{c}{\textbf{Places}} & \multicolumn{2}{c}{\textbf{Textures}} & & \\
 \multicolumn{1}{c}{} & FPR95 $\downarrow$  & AUROC $\uparrow$ & FPR95 $\downarrow$ & AUROC $\uparrow$ & FPR95 $\downarrow$ & AUROC $\uparrow$ & FPR95 $\downarrow$ & AUROC $\uparrow$ & FPR95 $\downarrow$ & AUROC $\uparrow$ \\
 \hline
MSP\cite{Kevin} & 54.99 & 87.74 & 70.83 & 80.86 & 73.99 & 79.76 & 68.00 & 79.61 & 66.95 & 81.99 \\
ODIN\cite{liang2018enhancing} & 47.66 & 89.66 & 60.15 & 84.59 & 67.89 & 81.78 & 50.23 & 85.62 & 56.48 & 85.41 \\
Mahalanobis\cite{lee2018simple} & 97.00 & 52.65 & 98.50 & 42.41 & 98.40 & 41.79 & 55.80 & 85.01 & 87.43 & 55.47 \\
Energy\cite{liu2020energy} & 55.72 & 89.95 & 59.26 & 85.89 & 64.92 & 82.86 & 53.72 & 85.99 & 58.41 & 86.17 \\
ReAct\cite{sun2021react} & 20.38 & 96.22 & 24.20 & 94.20 & 33.85 & 91.58 & 47.30 & 89.80 & 31.43 & 92.95 \\ 
DICE\cite{sun2022dice} & 25.63 & 94.49 & 35.15 & 90.83 & 46.49 & 87.48 & 31.72 & 90.30 & 34.75 & 90.77 \\
DICE + ReAct\cite{sun2022dice} & 18.64 & 96.24 & 25.45 & 93.94 & 36.86 & 90.67 & 28.07 & 92.74 & 27.25 & 93.40\\
\midrule
\textbf{LINe} \textbf{(Ours)} & \textbf{12.26} & \textbf{97.56} & \textbf{19.48} & \textbf{95.26} & \textbf{28.52} & \textbf{92.85} & \textbf{22.54} & \textbf{94.44} & \textbf{20.70} & \textbf{95.03}
\\
\bottomrule
\end{tabular}
}
\end{table*}

\noindent \textbf{Experimental Results.}
In Table \ref{tab:imagenet-results}, we reported the performances of four OOD test datasets respectively. 
The average results from the four OOD test datasets were also reported.
LINe outperformed all baselines including 
MSP\cite{Kevin},
ODIN\cite{liang2018enhancing}, 
Mahalanobis distance\cite{lee2018simple}, 
Energy score\cite{liu2020energy}, 
ReAct\cite{sun2021react}, 
DICE\cite{sun2022dice},
and DICE + ReAct\cite{sun2022dice}.
We compared LINe with Energy\cite{liu2020energy} first.
LINe drastically reduced the FPR95 by 37.71$\%$, which shows the benefit of leveraging important neurons under the same OOD scoring function.
Next, we compared LINe with ReAct\cite{sun2021react}.
LINe reduced the FPR95 by 10.73$\%$, which allows us to see the advantages of leveraging important neurons using Shapley value. LINe further outperformed recent DICE \cite{sun2022dice} and DICE + ReAct by 14.05\% and 6.55\%, respectively. Experimental results showed that the proposed method can be applied to real-world large dataset for OOD detection.

\subsection{Ablation Study}
In this section, we discuss the effectiveness of each part used in LINe  and the detailed differences from other similar approaches. 
We also analyze the effect of hyperparameters.

\subsubsection{Ablation Study of LINe on ImageNet}
Table \ref{tab:part-abl} shows an ablation study over various parts used in LINe.
As shown in the table, each part of Shapley-based pruning (AP and WP) improved the performance.
Comparing LINE w/o WP with Energy + AC allows us to see the advantages of reducing the noisy activation, which reduces FPR95 by 8.52$\%$.
Next, we compared LINe w/o AP with Energy + AC, which also shows the benefits of reducing noisy weights.
Compared to Energy + AC, LINe w/o AP reduced FPR95 by 12.21$\%$.
Finally, we compared LINe w/o AP with DICE + ReAct\cite{sun2022dice} at Table \ref{tab:imagenet-results}, which allows us to see the benefit of leveraging class-wise contribution under similar circumstances.
LINe w/o AP reduced the FPR95 by 4.06$\%$ from 27.25$\%$ to 23.19$\%$.
DICE + ReAct\cite{sun2022dice} uses activations to select important weights, while LINe w/o AP selects important weights using class-wise contribution derived from the Shapley value\cite{shapley1997value}.

\begin{table}[t]
\caption{\textbf{Ablation Study of LINe on ImageNet.} Ablation on the effectiveness of Shapley-based pruning used in LINe and comparison with similar approaches. Values are percentages and averaged over OOD datasets. AC, AP, and WP denote activation clipping, activation pruning, and weight pruning, respectively. 
}
\label{tab:part-abl}
\centering
\scalebox{0.85}{
    \begin{tabular}{lccccc}
    \toprule
    \textbf{Method} & \textbf{AC} & \textbf{AP} & \textbf{WP} & \textbf{FPR95}$\downarrow$ & \textbf{AUROC}$\uparrow$ \\ 
    \midrule
    Energy\cite{liu2020energy} &         &           &           & 58.41 & 86.17\\ 
    Energy + AC & \checkmark  &          &          & 35.40 & 91.86   \\
    LINe w/o WP &\checkmark  &\checkmark &           & 26.88 &   93.77  \\
    LINe w/o AP & \checkmark  &           &\checkmark & 23.19 &  94.57  \\
    \textbf{LINe (Ours)} &\checkmark  &\checkmark &\checkmark  & \textbf{20.70} & \textbf{95.03}   \\
    \bottomrule
    \end{tabular}
    }
\end{table}

\subsubsection{Effect of Changing AC Threshold on ImageNet}
In Section \ref{sec:AC}, we discussed the meaning of AC in terms of the neuron-concept association.
AC allows us to consider the number of activated high-level features in the penultimate layer.
In Table \ref{tab:clipping-results}, we show various OOD detection performances of the model by changing threshold $\delta$.
Starting from clipping threshold $\delta$ = $\infty$, the value of FPR95 is the highest in the table.
As clipping threshold $\delta$ becomes smaller, the OOD detection performance is improved.
But at clipping threshold $\delta <$ 0.8, the OOD detection performance dropped.
This is because, as the clipping threshold $\delta$ approaches 0, all the penultimate output values also approach zero. 
It is obvious that performance will drop when all the penultimate output values approach zero.

\begin{table}[t]
\caption{\small \textbf{Ablation on different thresholds ($\delta$) of AC.} Ablation on the different thresholds ($\delta$) of clipping. All values are percentages and averaged over multiple OOD test datasets. 
}
\label{tab:clipping-results}
\centering
\scalebox{0.85}{
        \begin{tabular}{lcc}
    \toprule
     \multicolumn{1}{l}{{\textbf{Threshold ($\delta$)}}} 
    & \multicolumn{1}{l}{\textbf{FPR95}}$\downarrow$ & \multicolumn{1}{l}{\textbf{AUROC}}$\uparrow$  \\
    \midrule
     $\delta = 0.1$ & 41.18 & 88.44 \\
     $\delta = 0.4$ & 23.43 & 94.79 \\
     $\delta = 0.8$ & \textbf{20.70} & \textbf{95.03} \\
     $\delta = 1.0$ & 21.69 & 94.81 \\
     $\delta = 1.5$ & 26.96 & 93.99 \\
     $\delta = 2.0$ & 31.88 & 92.97 \\
     $\delta = \infty$ (no AC) & 44.88 & 89.14 \\
    \bottomrule
    \end{tabular}
    }
\end{table}

\subsubsection{Effect of Changing Pruning Percentile on ImageNet}
\label{sec:pp-imagenet}
In this section, we conducted ablation studies on pruning percentiles ($p$) variation on ImageNet datasets as ID data.
In Table \ref{tab:imagep-results} we show effect of changing pruning percentile ($\mathit{p_{w}}$ and $\mathit{p_{a}}$) on ImageNet.
$\mathit{p_{a}}$ indicates pruning percentile for AP, $\mathit{p}_{w}$ indicates pruning percentile for WP.
For a fixed WP percentile $\mathit{p_{w}}$ with extremely high value (e.g., $\mathit{p_{w}}$ = 90), the performance tends to increase when the AP percentile $\mathit{p_{a}}$ falls.
Since LINe considers the number of activated features for detecting OOD samples, pruning most of the activations or weights has negatively impacted the performance. 
A lower pruning percentile is better to leverage differences in the number of activated features between ID and OOD samples.
But to restrict the negative effect of noisy signals, we need some portions that can remove the noisy signals.
As a result, the model performs the best when the pruning percentile is $\mathit{p_{w}}$ = 10 and $\mathit{p_{a}}$ = 10 on ImageNet.

\begin{table}[t]
\caption{\small \textbf{Effect of changing pruning percentile ($p$) on ImageNet.} Ablation on the different pruning percentile ($\mathit{p}_{a}$ and $\mathit{p}_{w}$) of pruning. All values are percentages and averaged over multiple OOD test datasets. $\mathit{p}_{a}$ denotes pruning percentile for AP, $\mathit{p}_{w}$ indicates pruning percentile for WP.}
\label{tab:imagep-results}
\centering
\scalebox{0.82}{
        \begin{tabular}{lccccc}
    \toprule
     & $\mathit{p}_a$ = 90 & $\mathit{p}_a$ = 70 & $\mathit{p}_a$ = 50 & $\mathit{p}_a$ = 30 & 
     $\mathit{p}_a$ = 10\\
     \cline{2-6}
  & \textbf{FPR95} $\downarrow$& \textbf{FPR95} $\downarrow$& \textbf{FPR95} $\downarrow$& \textbf{FPR95} $\downarrow$& \textbf{FPR95} $\downarrow$\\
    \midrule
     $\mathit{p_{w}}$ = 90 & 27.56 & 24.79 & 24.74 & 24.55 & 24.54 \\
     $\mathit{p_{w}}$ = 70 & 27.45 & 25.93 & 25.81 & 27.45 & 33.32 \\
     $\mathit{p_{w}}$ = 50 & 33.69 & 27.78 & 26.28 & 25.90 & 27.45 \\
     $\mathit{p_{w}}$ = 30 & 27.46 & 26.10 & 27.17 & 24.36 & 28.43 \\
     $\mathit{p_{w}}$ = 10 & 27.64 & 26.75 & 27.75 & 25.41 & \textbf{20.70} \\
    \bottomrule
    \end{tabular}
    }
\end{table}

\subsubsection{Effect of Changing Pruning Percentile on CIFAR Benchmarks}
In Table \ref{tab:cifar10p-results} and Table \ref{tab:cifar100p-results}, 
we show effect of changing pruning percentile ($\mathit{p_{w}}$ and $\mathit{p_{a}}$) on CIFAR Benchmarks.
Both tables show similar tendencies.
For all AP percentile $\mathit{p_{a}}$, the performance tends to increase when the WP percentile $\mathit{p_{w}}$ increases.
In Table \ref{tab:cifar10p-results}, highest performance appeared at $\mathit{p_{w}}$ = 90 and $\mathit{p_{a}}$ = 90.
On the other hand in Table \ref{tab:cifar100p-results}, the highest performance appeared at $\mathit{p_{w}}$ = 90 and $\mathit{p_{a}}$ = 10.
This result may seem to conflict with the result in Table \ref{tab:imagep-results}, our observation in Table \ref{tab:why-p} can explain the cause of the difference.

\begin{table}[t]
\caption{\small \textbf{Effect of changing pruning percentile ($p$) in CIFAR-10.} Ablation on the different pruning percentile ($\mathit{p}_{a}$ and $\mathit{p}_{w}$) of pruning. All values are percentages and averaged over multiple OOD test datasets. $\mathit{p}_{a}$ denotes pruning percentile for AP, $\mathit{p}_{w}$ indicates pruning percentile for WP.}
\label{tab:cifar10p-results}
\centering
\scalebox{0.82}{
        \begin{tabular}{lccccc}
    \toprule
     & $\mathit{p}_a$ = 90 & $\mathit{p}_a$ = 70 & $\mathit{p}_a$ = 50 & $\mathit{p}_a$ = 30 & 
     $\mathit{p}_a$ = 10\\
     \cline{2-6}
  & \textbf{FPR95} $\downarrow$& \textbf{FPR95} $\downarrow$& \textbf{FPR95} $\downarrow$& \textbf{FPR95} $\downarrow$& \textbf{FPR95} $\downarrow$\\
    \midrule
     $\mathit{p_{w}}$ = 90 & \textbf{14.72} & 15.00 & 15.00 & 15.00 & 14.99 \\
     $\mathit{p_{w}}$ = 70 & 14.80 & 15.12 & 15.12 & 15.12 & 15.10 \\
     $\mathit{p_{w}}$ = 50 & 14.80 & 15.12 & 15.11 & 15.11 & 15.10 \\
     $\mathit{p_{w}}$ = 30 & 14.80 & 15.12 & 15.11 & 15.12 & 15.10 \\
     $\mathit{p_{w}}$ = 10 & 14.80 & 15.13 & 15.13 & 15.12 & 15.73 \\
    \bottomrule
    \end{tabular}
    }
\end{table}

\begin{table}[t]
\caption{\small \textbf{Effect of changing pruning percentile ($p$) in CIFAR-100.} Ablation on the different pruning percentile ($\mathit{p}_{a}$ and $\mathit{p}_{w}$) of pruning. All values are percentages and averaged over multiple OOD test datasets. $\mathit{p}_{a}$ denotes pruning percentile for AP, $\mathit{p}_{w}$ indicates pruning percentile for WP.}
\label{tab:cifar100p-results}
\centering
\scalebox{0.82}{
        \begin{tabular}{lccccc}
    \toprule
     & $\mathit{p}_a$ = 90 & $\mathit{p}_a$ = 70 & $\mathit{p}_a$ = 50 & $\mathit{p}_a$ = 30 & 
     $\mathit{p}_a$ = 10\\
     \cline{2-6}
  & \textbf{FPR95} $\downarrow$& \textbf{FPR95} $\downarrow$& \textbf{FPR95} $\downarrow$& \textbf{FPR95} $\downarrow$& \textbf{FPR95} $\downarrow$\\
    \midrule
     $\mathit{p_{w}}$ = 90 & 38.75 & 37.81 & 37.81 & 37.75 & \textbf{35.67} \\
     $\mathit{p_{w}}$ = 70 & 38.37 & 39.30 & 40.07 & 39.75 & 40.81 \\
     $\mathit{p_{w}}$ = 50 & 38.37 & 39.19 & 40.54 & 40.27 & 42.14 \\
     $\mathit{p_{w}}$ = 30 & 38.37 & 39.19 & 40.65 & 40.21 & 39.32 \\
     $\mathit{p_{w}}$ = 10 & 38.40 & 39.31 & 40.76 & 40.91 & 38.17 \\
    \bottomrule
    \end{tabular}
    }
\end{table}

\subsubsection{Discussion}
\label{sec:disc}

\begin{table}[t]
\caption{\small \textbf{Percentage of class-specific neuron overlap in multiple classes.} Difference between the percentage of class-specific neuron overlap in multiple classes on three data sets. 
For each dataset, we calculated the proportion of very important (top 10$\%$) neurons in more than $o\%$ of the class. All values are percentages.
}
\label{tab:why-p}
\centering
\scalebox{0.82}{
        \begin{tabular}{lccc}
    \toprule
 \textbf{Overlap} & \textbf{CIFAR-10} & \textbf{CIFAR-100} & \textbf{ImageNet}\\
    \midrule
    $o$ = 20 & 24.56 & 26.90 & 1.70\\
    $o$ = 30 & 23.39 & 0.58 & 0.15\\
    \bottomrule
    \end{tabular}
    }
\end{table}
In Table \ref{tab:why-p}, we compared the percentage of class-specific neuron overlap in multiple classes on three data sets.
For each dataset, we calculated the proportion of very important (top 10$\%$) neurons in more than $o\%$ of the class.
These neurons activate in various classes which have semantically different features. 
Therefore, the higher proportion of these neurons can be seen as an overparameterized model with more numbers of generally activated neurons.
These overparameterized models make OOD detection difficult by creating noisy signals, which can be reduced by leveraging AP and WP.
To get optimal results from overparameterized models, we can lessen the effect of overparameterized weights with high WP percentile.
Also, the effectiveness of considering the number of activated class-specific neurons can be maximized in the low AP percentile as a tendency shown in Table \ref{tab:cifar100p-results}.
But for a much more overparameterized model, which is shown in Table \ref{tab:cifar10p-results}, we need to set high pruning percentile on both $\mathit{p_{w}}$ and $\mathit{p_{a}}$.

The degree of an overparameterized model can be understood from Section \ref{sec:imp-cifar}.
We used the same pre-trained model architecture for evaluating CIFAR-10 and CIFAR-100.
Of the two models with the same structure, we can intuitively understand that a model trained using a relatively small dataset is more likely to be overparameterized, which can also be observed in Table \ref{tab:why-p}.
The proportion of very important (top 10$\%$) neurons in more than 30$\%$ of the class(i.e., $o$ = 30) on CIFAR-10 is the largest compared to other datasets.
This observation shows that our pre-trained model used to evaluate CIFAR-10 is more overparameterized than other models we used.

\section{Conclusion}
\label{sec:Conclusion}

In this paper, we propose a powerful OOD Detection method called LINe. 
LINe adopts a neural-concept association, which only uses important activations and weights selectively by measuring class-wise contribution from the Shapley value.
Through LINe, we can effectively reduce the influence of the noise signal and make a difference in the overall OOD score between ID and OOD sample distribution.
We conducted extensive experiments to demonstrate that LINe is superior to state-of-the-art OOD detection methods and effective on multiple datasets.
From several theoretical studies and insights, we show how our method improves the performance of OOD detection.
Our method is effective but has some limitations.
It is fundamentally a trade-off relationship that pruning neurons to reduce the noisy output and considers the number of class-specific feature activations.
Users have to examine the trade-off considering the degree of overparameterization of the model.
We hope that as our study proposes an effective way to view OOD detection from a feature presentation perspective, attempts to understand the behavior of neural networks will be applied to multiple domains to discover other effective methods.
\section*{Acknowledgements}
This work was supported in part by the Institute of Information and Communications Technology Planning and Evaluation (IITP) grant funded by the Korea Government (MSIT)(No. 2022-0-00078: Explainable Logical Reasoning for Medical Knowledge Generation, No. 2021-0-02068: Artificial Intelligence Innovation Hub, No. RS-2022-00155911: Artificial Intelligence Convergence Innovation Human Resources Development (Kyung Hee University)) and by the National Research Foundation of Korea (NRF) grant funded by the Korea government(MSIT) (No. 2021R1G1A1094990).

{\small
\bibliographystyle{ieee_fullname}

}

\clearpage
\onecolumn

\begin{center}
{\Large \bf Supplementary Material for LINe: \\Out-of-Distribution Detection by Leveraging Important Neurons}
\end{center}

\vspace{1em}

\appendix

\section{Detailed CIFAR Benchmark Results}
Table \ref{tab:cifar10-detail-results} and Table \ref{tab:cifar100-detail-results} are detailed results of CIFAR-10 and CIFAR-100 benchmark experiments (detailed results for Table 1 in the main text).
For both tables, 
all the results except DICE + ReAct and LINe are taken from Sun et al. \cite{sun2022dice}.
We choose hyperparameters for DICE + ReAct as sparsity $p$ = 90 and ReAct threshold = 1.0, as in \cite{sun2022dice,sun2021react}. 

\section{LINe on Other Models}
In this section, we show LINe also works well with other models.
In the main text, we show LINe with pre-trained DenseNet\cite{huang2017densely} and ResNet-50\cite{he2016identity} on CIFAR and ImageNet datasets, respectively.
In this section, we show LINe can be used for MobileNetV2~\cite{sandler2018mobilenetv2}, which is pre-trained on the ImageNet-1k dataset from PyTorch.
Experiment settings are the same in Section 4.2.
We choose hyperparameters for LINe as pruning percentile $p_{w} = p_{a} = 10$ and clipping threshold $\delta = 0.6$.
As shown in Table \ref{tab:imagenet-mobile}, our method implemented on MobileNetV2 outperformed all the other methods.

\section{LINe with Other Shapley-value Approximation}
We use the Taylor approximation in the main text to compute the Shapley value.
To see the difference of changing approximation to compute the Shapley value, we use IntGrad approximation, which is also introduced in \cite{khakzar2021neural}.
For input $x^{l} \in \mathit{D}$, where $x^{l}$ denotes the sample of class $l$ from dataset $\mathit{D}$, a contribution(i.e., Shapley value) of $i$-th neuron $a_{i}$ in class $l$, $s_{i}^{l}$ is calculated as 
\begin{equation}\label{equ:intgrad}
    s_{i}^{l} = a_{i}^{l}\int_{\alpha = 0}^{1}\frac{\partial f_{\theta}(\alpha a_{i}^{l};x)}{\partial a_{i}^{l}}d\alpha.
\end{equation}
Contribution matrix $C_{int}$ can be defined with contribution calculated by Equation \ref{equ:intgrad}.
With this contribution matrix $C_{int}$, we can apply LINe.
Table \ref{tab:shap-compare} show the result of LINe with Taylor and IntGrad approximation.
The results of both methods are the same.
Calculated contributions from both methods are different, but the order of \emph{top-k} neurons is still the same.
However, the precomputing time of IntGrad is almost 11 times larger than the Taylor approximation, so it is better to choose Taylor as an approximation method.

\section{Additional Theoretical Analysis}
The outstanding performance of LINe is grounded on three different groups of papers in the related work section (Sec 2.1-2.3).
In Network Dissection \cite{bau2017dissection} and HINT \cite{wang2022hint}, neurons in the deep layer (e.g., penultimate layer) represent a specific concept (e.g., window, mammal).
Also, in Khazar et al. \cite{khakzar2021neural}, neurons with high Shapely values have critical fragments of the encoded input information.
We draw an insight from the above studies that a group of neurons in the penultimate layer with high Shapley values for a specific class has essential concepts for classifying that class.
We call this group of neurons class-specific neurons.
Therefore, we can select important class-specific neurons and mask less important neurons by ranking the contribution of neurons.
The pruning parts in LINe (i.e., AP and WP) improve the performance by masking less important neurons which trigger noisy outputs.
Since class-specific neurons are activated only for essential concepts for each class, OOD samples with different visual features (i.e., concept) cannot activate most of the class-specific neurons.
This simple idea motivates AC by limiting the size of activation, which makes AC treat class-specific features equally and improves OOD detection performance.

\begin{table*}[htb!]
\centering
\caption{\textbf{Comparison on CIFAR-10 benchmark.} Table shows comparison with competitive post-hoc OOD detection methods on CIFAR-10 benchmark. All values in this table are percentages. The average over six OOD test datasets is also reported.
}
\label{tab:cifar10-detail-results}
\scalebox{0.72}{
\begin{tabular}{lcccccccccccccc}
    \toprule
\multirow{3}{*}{\textbf{Method}}  & \multicolumn{12}{c}{\textbf{OOD Datasets}} & \multicolumn{2}{c}{\multirow{2}{*}{\textbf{Average}}} \\ \cline{2-13}
 \multicolumn{1}{c}{} &\multicolumn{2}{c}{\textbf{SVHN}} & \multicolumn{2}{c}{\textbf{Textures}} & \multicolumn{2}{c}{\textbf{iSUN}} & \multicolumn{2}{c}{\textbf{LSUN}} & \multicolumn{2}{c}{\textbf{LSUN-Crop}} & \multicolumn{2}{c}{\textbf{Places365}} & &\\
 \multicolumn{1}{c}{} & FPR95  & AUROC & FPR95  & AUROC  & FPR95 & AUROC  & FPR95 & AUROC  & FPR95  & AUROC  & FPR95 & AUROC& FPR95 & AUROC  \\
 \hline
MSP\cite{Kevin}& 47.24 & 93.48 & 64.15 & 88.15 & 42.31 & 94.52 & 42.10 & 94.51 & 33.57 & 95.54 & 63.02 & 88.57 & 48.73 & 92.46 \\
ODIN\cite{liang2018enhancing} & 25.29 & 94.57 & 57.50 & 82.38 & 3.98 & 98.90 & 3.09 & 99.02 & 4.70 & 98.86 & 52.85 & 88.55 & 24.57 & 93.71 \\
Mahalanobis\cite{lee2018simple}& 6.42 & 98.31 & 21.51 & 92.15 & 9.78 & 97.25 & 9.14 & 97.09 & 56.55 & 86.96 & 85.14 & 63.15 & 31.42 & 89.15\\
Energy\cite{liu2020energy}& 40.61 & 93.99 & 56.12 & 86.43 & 10.07 & 98.07 & 9.28 & 98.12 & 3.81 & 99.15 & 39.40 & 91.64 & 26.55 & 94.57\\
ReAct\cite{sun2021react}& 41.64 & 93.87 & 43.58 & 92.47 & 12.72 & 97.72 & 11.46 & 97.87 & 5.96 & 98.84 & 43.31 & 91.03 & 26.45 & 94.67 \\ 
DICE\cite{sun2022dice}& 25.99 & 95.90 & 41.90 & 88.18 & 4.36 & 99.14 & 3.91 & 99.20 & 0.26 & 99.92 & 48.59 & 89.13 & 20.83 & 95.24\\
DICE + ReAct\cite{sun2022dice}& 12.49 & 97.61 & 25.83 & 94.56 & 5.27 & 99.02 & 3.95 & 99.14 & 0.43 & 99.89 & 50.94 & 89.63 & 16.48 & 96.64\\
\midrule
\textbf{LINe} \textbf{(Ours)} & 11.38 & 97.75 & 23.44 & 95.12 & 4.90 & 99.01 & 4.19 & 99.09 & 0.61 & 99.83 & 43.78 & 91.12 & 14.72 & 96.99\\
\bottomrule
\end{tabular}
}
\vspace*{6mm}
\end{table*}

\begin{table*}[htb!]
\centering
\caption{\textbf{Comparison on CIFAR-100 benchmark.} Table shows comparison with competitive post-hoc OOD detection methods on CIFAR-100 benchmark. All values in this table are percentages. The average over six OOD test datasets is also reported.
}
\label{tab:cifar100-detail-results}
\scalebox{0.72}{
\begin{tabular}{lcccccccccccccc}
    \toprule
\multirow{3}{*}{\textbf{Method}}  & \multicolumn{12}{c}{\textbf{OOD Datasets}} & \multicolumn{2}{c}{\multirow{2}{*}{\textbf{Average}}} \\ \cline{2-13}
 \multicolumn{1}{c}{} &\multicolumn{2}{c}{\textbf{SVHN}} & \multicolumn{2}{c}{\textbf{Textures}} & \multicolumn{2}{c}{\textbf{iSUN}} & \multicolumn{2}{c}{\textbf{LSUN}} & \multicolumn{2}{c}{\textbf{LSUN-Crop}} & \multicolumn{2}{c}{\textbf{Places365}} & &\\
 \multicolumn{1}{c}{} & FPR95  & AUROC & FPR95  & AUROC  & FPR95 & AUROC  & FPR95 & AUROC  & FPR95  & AUROC  & FPR95 & AUROC& FPR95  & AUROC  \\
 \hline
MSP\cite{Kevin}& 81.70 & 75.40 & 84.79 & 71.48 & 85.99 & 70.17 & 85.24 & 69.18 & 60.49 & 85.60 & 82.55 & 74.31 & 80.13 & 74.36 \\
ODIN\cite{liang2018enhancing} & 41.35 & 92.65 & 82.34 & 71.48 & 67.05 & 83.84 & 65.22 & 84.22 & 10.54 & 97.93 & 82.32 & 76.84 & 58.14 & 84.49\\
Mahalanobis\cite{lee2018simple}& 22.44 & 95.67 & 62.39 & 79.39 & 31.38 & 93.21 & 23.07 & 94.20 & 68.90 & 86.30 & 92.66 & 61.39 & 55.37 & 82.73\\
Energy\cite{liu2020energy}& 87.46 & 81.85 & 84.15 & 71.03 & 74.54 & 78.95 &  70.65 & 80.14 &  14.72 & 97.43 & 79.20 & 77.72 & 68.45 & 81.19\\
ReAct\cite{sun2021react}& 83.81 & 81.41 & 77.78 & 78.95 & 65.27 & 86.55 & 60.08 & 87.88 & 25.55 & 94.92 &  82.65 & 74.04 & 62.27 & 84.47\\
DICE\cite{sun2022dice}& 54.65 & 88.84 & 65.04 & 76.42 & 48.72 & 90.08 & 49.40 & 91.04 & 0.93 & 99.74 & 79.58 & 77.26 & 49.72 & 87.23 \\
DICE + ReAct\cite{sun2022dice}& 55.52 & 88.02 & 41.54 & 86.26 & 44.32 & 91.44 & 54.44 & 89.84 & 7.56 & 98.61 & 94.05 & 56.26 & 49.57 & 85.07\\
\midrule
\textbf{LINe} \textbf{(Ours)} & 31.10 & 91.90 & 39.29 & 87.84 & 24.07 & 94.85 & 25.32 & 94.63 & 5.72 & 98.87 & 88.50 & 63.93 & 35.67 & 88.67
\\
\bottomrule
\end{tabular}
}
\vspace*{6mm}
\end{table*}

\begin{table*}[htb!]
\centering
\caption{\textbf{LINe with MobileNetV2 on ImageNet benchmark.} Results compared with competitive post-hoc OOD detection methods on ImageNet benchmark are reported. All values in this table are percentages and averaged over four OOD test datasets.
$\downarrow$ indicates smaller value means higher performance and $\uparrow$ indicates vice versa.
} 
\label{tab:imagenet-mobile}
\scalebox{0.83}{
\begin{tabular}{lcccccccccc}
    \toprule
  \multirow{3}{*}{\textbf{Method}}  & \multicolumn{8}{c}{\textbf{OOD Datasets}} & \multicolumn{2}{c}{\multirow{2}{*}{\textbf{Average}}} \\ \cline{2-9}
 \multicolumn{1}{c}{} & \multicolumn{2}{c}{\textbf{iNaturalist}} & \multicolumn{2}{c}{\textbf{SUN}} & \multicolumn{2}{c}{\textbf{Places}} & \multicolumn{2}{c}{\textbf{Textures}} & & \\
 \multicolumn{1}{c}{} & FPR95 $\downarrow$  & AUROC $\uparrow$ & FPR95 $\downarrow$ & AUROC $\uparrow$ & FPR95 $\downarrow$ & AUROC $\uparrow$ & FPR95 $\downarrow$ & AUROC $\uparrow$ & FPR95 $\downarrow$ & AUROC $\uparrow$ \\
 \hline
MSP\cite{Kevin} & 64.29 & 85.32 & 77.02 & 77.10 & 79.23 & 76.27 & 73.51 & 77.30 & 73.51 & 79.00 \\
ODIN\cite{liang2018enhancing} & 55.39 & 87.62 & 54.07 & 85.88 & 57.36 & 84.71 & 49.96 & 85.03 & 54.20 & 85.81 \\
Mahalanobis\cite{lee2018simple} & 62.11 & 81.00 & 47.82 & 86.33 & 52.09 & 83.63 & 92.38 & 33.06 & 63.60 & 71.01 \\
Energy score\cite{liu2020energy} & 59.50 & 88.91 & 62.65 & 84.50 & 69.37 & 81.19 & 58.05 & 85.03 & 62.39 & 84.91 \\
ReAct~\cite{sun2021react} & 42.40 & 91.53 & 47.69 & 88.16 & 51.56 & 86.64 & 38.42 & 91.53 & 45.02 & 89.47 \\
DICE~\cite{sun2022dice} & 43.09 &  90.83 & 38.69 & 90.46 & 53.11 & 85.81 & 32.80 & 91.30 & 41.92 & 89.60 \\
DICE + ReAct~\cite{sun2022dice} & 32.30 & 93.57 & 31.22 & 92.86 & 46.78 & 88.02 & 16.28 & 96.25 & 31.64 & 92.68 \\
\midrule
\textbf{LINe} \textbf{(Ours)} & 24.95 & 95.53 & 33.19 & 92.94 & 47.95 & 88.98 & 12.30 & 97.05 & 29.60 & 93.62
\\
\bottomrule
\end{tabular}
}
\vspace*{6mm}
\end{table*}

\begin{table*}[htb!]
\caption{\textbf{Comparison on different approximation methods.} Results with different Shapley value approximation are reported on CIFAR-10, CIFAR-100, ImageNet benchmarks. All values in this table are percentages and averaged.
$\downarrow$ indicates smaller value means better performance and $\uparrow$ indicates vice versa. 
}
\centering
\scalebox{0.8}{
\begin{tabular}{lcc|cc|cc}
\toprule
\multirow{2}{*}{\textbf{Method}} & \multicolumn{2}{c|}{\textbf{CIFAR-10}} & \multicolumn{2}{c|}{\textbf{CIFAR-100}} & \multicolumn{2}{c}{\textbf{ImageNet}}\\
 \multicolumn{1}{c}{} & \multicolumn{1}{c}{\textbf{FPR95 $\downarrow$}} & \multicolumn{1}{c|}{\textbf{AUROC $\uparrow$}} & \multicolumn{1}{c}{\textbf{FPR95 $\downarrow$}} & \multicolumn{1}{c|}{\textbf{AUROC $\uparrow$}} & \multicolumn{1}{c}{\textbf{FPR95 $\downarrow$}} & \multicolumn{1}{c}{\textbf{AUROC $\uparrow$}}\\
 \midrule
 Taylor  & 14.71 & 96.99 & 35.67 & 88.67 & 20.70 & 95.03\\
 IntGrad  & 14.71 & 96.99 & 35.67 & 88.67 & 20.70 & 95.03\\
\bottomrule
\end{tabular}}
\label{tab:shap-compare}
\vspace*{6mm}
\end{table*}

\end{document}